\documentclass{article}

\usepackage{wrapfig} 
\usepackage[preprint]{neurips_2022}

\usepackage[utf8]{inputenc} 
\usepackage[T1]{fontenc}    
\usepackage{hyperref}       
\usepackage{url}            
\usepackage{booktabs}       
\usepackage{amsfonts}       
\usepackage{nicefrac}       
\usepackage{microtype}      
\usepackage{xcolor}         
\usepackage{amsmath}
\usepackage{graphicx}
\usepackage[ruled]{algorithm2e}
\usepackage{multirow}
\usepackage{makecell}
\PassOptionsToPackage{numbers,author,year,sort&compress,round}{natbib}
\newcommand{\citen}[1]{[\citenum{#1}]} 
\newcommand{\citef}[1]{\citeauthor{#1} [\citenum{#1}]} 

\newcommand{\cut}[1]{}

\title{Nuclear Norm Maximization Based Curiosity-Driven Learning}

\author{%
Chao Chen\textsuperscript{1}, Zijian Gao\textsuperscript{1}, Kele Xu\textsuperscript{1,}\thanks{Corresponding author} ,
 Sen Yang\textsuperscript{1}, Yiying Li\textsuperscript{2}, \\ \textbf{Bo Ding\textsuperscript{1}, Dawei Feng\textsuperscript{1}, Huaimin Wang\textsuperscript{1}} \\
\textsuperscript{1} National University of Defense Technology, Changsha, China\\
\textsuperscript{2} Artificial Intelligence Research Center, DII, Beijing, China\\
\texttt{kelele.xu@gmail.com} \\
}

\begin{document}

\maketitle

\begin{abstract}
To handle the sparsity of the extrinsic rewards in reinforcement learning, researchers have proposed intrinsic reward which enables the agent to learn the skills that might come in handy for pursuing the rewards in the future, such as encouraging the agent to visit novel states. However, the intrinsic reward can be noisy due to the undesirable environment's stochasticity and directly applying the noisy value predictions to supervise the policy is detrimental to improve the learning performance and efficiency. Moreover, many previous studies employ $\ell^2$ norm or variance to measure the exploration novelty, which will amplify the noise due to the square operation. In this paper, we address aforementioned challenges by proposing a novel curiosity leveraging the nuclear norm maximization (NNM), which can quantify the novelty of exploring the environment more accurately while providing high-tolerance to the noise and outliers. We conduct extensive experiments across a variety of benchmark environments and the results suggest that NNM can provide state-of-the-art performance compared with previous curiosity methods. On 26 Atari games subset, NNM achieves a human-normalized score of 1.09, which doubles that of competitive intrinsic rewards-based approaches. Our code will be released publicly to enhance the reproducibility.
\end{abstract}

\section{Introduction}
In the past few years, deep reinforcement learning (DRL) has achieved spectacular performance on a wide range of challenging sequential decision-making tasks, ranging from Atari games \citen{dqn}, the board game Go \citen{silver2017mastering}, StarCraft game \citen{vinyals2019grandmaster}, to multiple robotic control tasks \citen{yang2018hierarchical}.
Despite these promising results, the DRL systems rely on well-defined extrinsic reward from the environment, which requires significant human engineering. It is impractical to manually engineer dense reward functions for every task which an agent aims to solve \citen{RND}. Moreover, the extrinsic rewards of real-world tasks can be sparse or missing. Many of the previous DRL methods struggle with the heavy design of the extrinsic reward and the sparsity of the external reward.

As human agents, we are driven by the curiosity to learn new skills which might come in handy for pursuing the rewards in the future, and are accustomed to handle the sparse external rewards \citen{ICM}. Previous findings from the developmental psychologists also suggest that intrinsic motivation (i.e., curiosity) may be the primary driver for the development of children \citen{burda2018large,ryan2000intrinsic,smith2005development}. Motivated by human intrinsic motivation, a few attempts employ the intrinsic reward to improve the performance of DRL by generating curiosity during the learning phase ~\citen{liu2021behavior,RND,Disagreemet}.
Most formulations of curiosity can be grouped into two broad classes \citen{ICM}: novelty-search based and prediction-error based methods. Novelty-search based approaches assign the intrinsic rewards when the agent moves to a new state \citen{brafman2002r,kearns2002near}. Prediction-error based approaches encourage the agent to perform actions which can reduce the prediction error based on its knowledge about the environment~\citen{ICM,schmidhuber1991possibility,chentanez2004intrinsically,stadie2015incentivizing}. Previous studies~\citen{Disagreemet,liu2021behavior,ICM,RND,tao2020novelty} demonstrate that intrinsic rewards could be helpful to alleviate the problems of sparse extrinsic rewards. 

\begin{table}[!htbp]
    \centering
    \caption{\footnotesize{The calculations of the curiosity-based intrinsic rewards. The encoded state-space and action can be denoted as $\mathbf{z}_t$ and $\mathbf{a}_t$ in time-step $t$, $g(\cdot)$ is the prediction model, $c$ is a positive constant value.}}
    \label{tab:calculate-of-int-reward}
    \begin{tabular}{lll}
     \toprule
     Method    & Calculation & Description \\
     \midrule  
     ICM \citen{ICM}                             & $\|g(\mathbf{z}_{t+1}|\mathbf{z}_t,\mathbf{a}_t )-\mathbf{z}_{t+1}\|_2^2$ 
     & $g(\cdot)$ is the predict model \\
     Disagreement\citen{Disagreemet}            & $Variance\{g_i(\mathbf{z}_{t+1}|\mathbf{z}_t,\mathbf{a}_t)\}, \quad i=1,\cdots,n$ 
     & n predict models\\
     RND \citen{RND}                             & $\|g(\mathbf{z}_t,\mathbf{a}_t )-\hat{g}(\mathbf{z}_t,\mathbf{a}_t)\|_2^2$ 
     & $\hat{g}$ means frozen network \\
     \midrule
     APT \citen{liu2021behavior}                             & $\textstyle{\sum_{j\in random}\log \|\mathbf{z}_t-\mathbf{z}_j\|_2^2} \quad j=1,\cdots,k$ &k-nearest neighbors \\
     ProtoRL \citen{ProtoRL}                     & $\textstyle{\sum_{j\in prototypes}\log \|\mathbf{z}_t-\mathbf{z}_j\|_2^2} \quad j=1,\cdots,k$ & k-nearest neighbors \\
     \midrule
     NGU \citen{NGU}                             & $\textstyle{ 1/(\sqrt{\sum_{f_i \in N_k}K(\mathbf{z}_t,\mathbf{z}_i)}}+c)$   & \makecell[l]{$K(\cdot)$ is inverse to\\  Euclidean distance} \\  
     \bottomrule
    \end{tabular}
\end{table}
\par
Despite the success of curiosity-based approaches in DRL, current approaches are still confronted with several challenges. Firstly, in DRL, the deep model is used to generate value predictions to supervise the learning. Due to the contamination of the noise, such as the environment stochasticity, the value predictions can be inaccurate while the policy networks fail to model the uncertainty of their predictions or actions. Directly applying the noisy values to supervise the learning can be detrimental to the performance \citen{mai2022sample}, as these value predictions are unstable. Very few attempts have been made to mitigate the impact of undesirable noise. Secondly, many of previous attempts employed the $\ell^2$ norm or the variance (as shown in Table \ref{tab:calculate-of-int-reward}) to measure the novelty. It is widely-known that both $\ell^2$ norm and variance are sensitive to the outliers and the noise, especially for the high-dimensional states, as the impact of the noisy values can be amplified due to the squared terms. 


\par
\begin{figure*}[!htb]  
    \centerline{\includegraphics[width=0.6\textwidth]{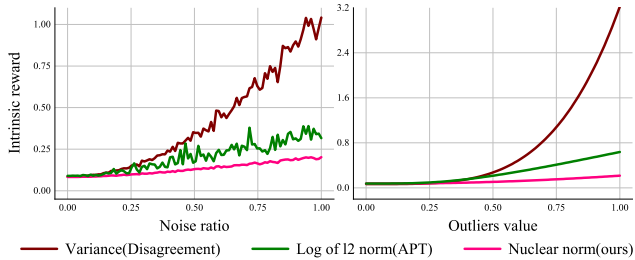}}
    \caption{\footnotesize{Calculated intrinsic reward when the observation input $\mathbf{Z}$ is contaminated by noise or outliers. $\mathbf{Z}$ is a $5 \times 128$ matrix (with 5 states, 128 dimensions). Left: $\mathbf{Z}$ is with random noise sampled from $\mathcal{N}(0, \epsilon^2)$; Right: $\mathbf{Z}$ is with the synthetic outlier (add a random value to $\mathbf{Z}$). Curves show the produced intrinsic reward results when $\epsilon$ and outliers gradually increase from 0 to 1.}}
    \label{fig-compare-of-norm}
\end{figure*}
\par
We address aforementioned challenges by proposing a novel intrinsic reward, leveraging the nuclear norm maximization (NNM). As Figure \ref{fig-compare-of-norm} shows, a toy experiment is conducted using the synthetic data to demonstrate the advantages of nuclear norm-based approach \cut{compared with previous methods, from the perspectives of }as to the robustness to the noise and outliers. Here, we compare the nuclear norm with two representative curiosity \cut{approaches}forms: the variance and $\ell^2$ norm (in its logarithmic form), which correspond to Disagreement \citen{Disagreemet} and APT \citen{liu2021behavior} respectively (introduced in Table \ref{tab:calculate-of-int-reward}).
We provide the calculated intrinsic rewards of different forms in Figure \ref{fig-compare-of-norm} with random noise and outliers added. We see that the nuclear norm is much more stable and robust to the noise and outliers changes of input.

\par
To summarize, our main contributions are: (1) We firstly analyze the shortcomings of previous curiosity-based intrinsic rewards, especially the lack of robustness to the noise and outliers.
(2) We present a new intrinsic reward based on the NNM, which can actively encourage the exploration of novel state in the task-agnostic environment. Our solution can provide superior performance, while retaining higher noise or outlier tolerance.
Moreover, our method is easy to implement and adds minimal overhead to the computation performed. (3) We show that our method significantly outperforms the competitive intrinsic rewards on multiple benchmarks including DeepMind Control Suite (DMC) \citen{dm_control} and Atari Suite \citen{atari}, either training the agent with only intrinsic reward or combined with extrinsic rewards. On 26 Atari games subset, NNM archives a human-normalized score of 1.09, which doubles that of competitive intrinsic rewards-based approaches. On DMC suites, we achieve state-of-the-art performance on 11 of 12 tasks.

\section{Related work}{\label{Related_work}}
\textbf{Curiosity}: Many attempts adopt the framework of intrinsic reward $r^{int}$ to improve the performance of DRL. One representative category of intrinsic rewards is called curiosity, which endeavors to make agents curious to unknown states and focuses on novelty search.
One intuitive curiosity is the count-based approach, which is not scalable in high-dimensional situations.
DDQN-PC~\citen{bellemare2016unifying}, A3C+~\citen{bellemare2016unifying}, DQN-PixelCNN~\citen{ostrovski2017count}, and $\phi$-EB~\citen{10.5555/3172077.3172232} adopt the density models to measure the visited time of states, which allows to compute the pseudo-count. Inverse Curiosity Model (ICM)~\citen{ICM} is one pioneer work for curiosity. ICM trains a prediction model which takes current feature state and current action as input, and outputs a next predicted state, the distance between predicted state and true state is applied as intrinsic reward, with the intuition that more visited state-action pairs as training data would lead to more accurate prediction. Similar to ICM~\citen{ICM}, RND~\citen{RND} assesses the state novelty by distilling a random neural network, which is simpler in structure. However, these approaches suffer from the undesirable-noise issue~\citen{ICM}. Disagreement~\citen{Disagreemet} learns the ensemble of prediction models instead of a single model, and explores the variance of different abstract predicted representations to measure the novelty. As the state-of-the-art approach using the pre-training and fine-tune paradigm, Unsupervised Active Pre-training (APT) \citen{liu2021behavior} aims to motivate the agent with a reward $r_t$ to maximize the entropy in an abstract representation space. Either the variance or the $\ell^2$ norm can amplify the noise due to the square operations. Very few attempts have been made to mitigate the noise for the intrinsic rewards, which however is the main goal of this paper. We explore a novel intrinsic reward leveraging the nuclear norm.
\par
\textbf{Nuclear Norm}: 
As one popular convex surrogate function for the $\operatorname{rank}$ function,
nuclear norm is widely-used for low-rank modeling. Low-rank matrices appear throughout sciences and engineering, such as in computer vision, recommendation systems, bioinformatics and economics~\citen{li2020empirical,udell2019big}. Due to the incomplete or corrupted observation of the low-rank matrix, many of previous attempts focus on the low-rank matrix recovery problem \citen{fan2019factor}. \citeauthor{hu2021low} [\citenum{hu2021low}] provides a comprehensive survey for Low Rank Regularization (LRR), mainly analysing the effect of using LRR as loss function by nuclear norm or other tools. 
\citef{cui2020towards} and \citef{xiong2021wrmatch} apply nuclear norm and weighted nuclear norm as loss functions to minimize $\operatorname{rank}$ of matrix for image classification task respectively, and both of them achieve superior results. 
However, our work aims to maximize the nuclear norm of the encoded states induced by the policy. By utilizing the properties of nuclear norm,
our proposed method is more robust to the undesirable noise, and performs better or comparably well as other more complex and specialized state-of-the-art methods.




\section{Nuclear Norm Maximization-based Curiosity}{\label{methods}} 
Our method to improve curiosity of the agent is based on the nuclear norm. We will first introduce the nuclear norm and notation briefly, then describe our NNM approach. 


\subsection{Nuclear Norm}
For a vector $\mathbf{z} \in \mathbb{R}^{1 \times m}$, the $\ell^{1}$ and $\ell^{2}$ norms can be denoted as $\|\mathbf{z}\|_{1}=\sum_{i=1}^{m}\left|z_{i}\right|$ and $\|\mathbf{z}\|_{2}=\sqrt{\sum_{i=1}^{m} z_{i}^{2}}$ respectively. Let $\mathbf{Z}=\left[\mathbf{z}_{1},\mathbf{z}_{2}, \ldots, \mathbf{z}_{n}\right]$ $\in \mathbb{R}^{m \times n}$, where the $i$-th column of $\mathbf{Z}$ is denoted as $\mathbf{z}_{i}=\left(z_{i 1}, z_{i 2}, \ldots\right.$, $\left.z_{i j}, \ldots, z_{i m}\right) \in \mathbb{R}^{1 \times m}$. The Frobenius norm ($F$-norm) of the matrix $\mathbf{Z}$ is defined as $\|\mathbf{Z}\|_{F}=\sqrt{\sum_{i=1}^{n}\left\|\mathbf{z}_{i}\right\|_{2}^{2}}$, which denotes the sum of the squares of all elements in the matrix and then calculate the square root. 
Suppose $\mathbf{Z}$ is a linear independent matrix, the singular values of $\mathbf{Z}$ can be denoted as $\sigma_i, \left( i=1,2,\cdots,d\right), d=\min \left(m,n \right)$. Then the eigenvalues of matrix $\mathbf{Z}^\top \mathbf{Z}$ are $\sigma_i^2, \left( i=1,2,\cdots,d\right)$ .
The nuclear norm of $\mathbf{Z}$ can be denoted as:
\begin{equation}
   \|\mathbf{Z}\|_{*}= \operatorname{trace}\left(\left(\mathbf{Z}^\top \mathbf{Z}\right)^{1 / 2}\right)=\sum_{i=1}^{d} \sigma_{i} ,
\end{equation}
which constitutes a special case of the Schatten norm $\|\mathbf{Z}\|_{p}=\left(\sum_{i=1}^{d} \sigma_{i}^{p}\right)^{\frac{1}{p}}$. 
Nuclear Norm is also known as trace norm, or schatten $1$ norm. 
Due to the incomplete or corrupted observations, the matrix can be contaminated. 
Thus, the recovery of data from the corrupted data has become increasingly important. In most cases of interest, this inverse problem can be severely ill-posed and the accurate estimation heavily relies on some prior knowledge. For example, previous attempts employ the inherent redundancy of the underlying modeling matrix, which can be further formulated as a matrix completion problem using the low-rank regularization \citen{morison2021sure}. 
The low-rank regularized estimation issue is formulated as: $\underset{\mathbf{Z}}{\text{min}} \operatorname{rank}(\mathbf{Z})$.

\subsection{Nuclear-Norm Maximization-based Intrinsic Rewards}

\begin{figure*}[!htb]
\centerline{\includegraphics[width=0.6\textwidth]{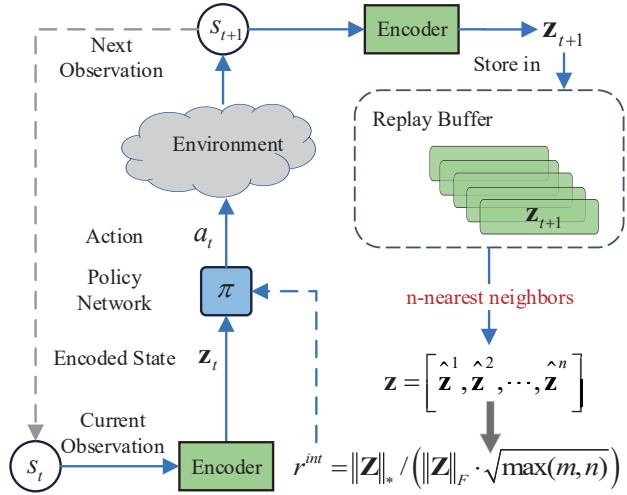}}
\caption{\footnotesize{{\bfseries Overview structure of NNM:} At time step $t$, the agent in the state $s_t$ takes action $a_t$ generated by policy $\pi$. Then the agent interacts with the environment and gets the next state $s_{t+1}$. We concatenate the $n$ encoded states as matrix $\mathbf{Z}$, and calculate the nuclear norm $\left\| \mathbf{Z} \right\|_\ast$ as intrinsic reward $r_t^{int}$ to train policy $\pi$}. 
}  
    \label{fig-whole-frame} 
\end{figure*}
Our main idea is to accurately quality novelty by nuclear norm of matrix when visiting a state, alleviating the perturbation of noise and outliers. Inspired by Disagreement \citen{Disagreemet} and APT \citen{liu2021behavior}, the matrix was generated by the ensemble of prediction models or stored memory. Concretely, the matrix with the size of $m\times n$ consists of $n$ encoded states and each state is $m$-dimension, which come from outputs $\hat{\mathbf{z}}^1,\hat{\mathbf{z}}^2,\cdots,\hat{\mathbf{z}}^n$ of $n$ prediction models, or spliced by current state $\hat{\mathbf{z}}^1$ and its $(n-1)$ nearest neighbor states $\hat{\mathbf{z}}^2,\hat{\mathbf{z}}^3,\cdots,\hat{\mathbf{z}}^n$. Here, $\mathbf{z}$ represents the abstract state\cut{-space} by mapping raw high-dimensional observation to low-dimensional abstract space. 
As each column of the matrix $\mathbf{Z}$ represents an encoded state, the $\operatorname{rank}(\mathbf{Z})$ can be used to represent the diversity within the matrix, due to the fact that higher $\operatorname{rank}(\mathbf{Z})$ denotes larger linear irrelevance among the encoded states.
Unlike previous\cut{attempts} studies and applications based on the rank of matrix \citen{fazel2002matrix, cui2020towards, hu2021low, xiong2021wrmatch} employing the low-rank modeling, on the contrary, we \cut{can} creatively use the rank by maximizing $\operatorname{rank}(\mathbf{Z})$ to enlarge the exploration diversity, which encourages the agent to visit more different states of high diversity. Thus, the intuition of our intrinsic reward can be achieved by: $\underset{\mathbf{Z}}{\text{max}} \operatorname{rank}(\mathbf{Z})$.

Generally, two ways can be used to $\underset{\mathbf{Z}}{\text{max}} \operatorname{rank}(\mathbf{Z})$: as loss function or as reward. However, directly maximizing the matrix rank is an NP-hard non-convex problem so it can't be taken as loss function. Besides, the value of matrix rank is discrete and can not precisely reflect the novelty of states, so applying raw value of matrix rank as reward to guide the learning wouldn't achieve our goal well.
Mathematically, just as $\ell^1$ norm is the tightest convex relaxation of $\ell^0$ norm for a vector, the calculation of matrix rank is usually replaced by the nuclear-norm, which have been proved the convex envelope of $\operatorname{rank}$ \citen{fazel2002matrix}. Therefore, the novelty could be maintained by maximizing nuclear norm approximately.
Compared with $\operatorname{rank}$, the nuclear norm has several good properties: Firstly, the convexity of the nuclear norm makes it possible to develop fast and convergent algorithms in optimization \citen{zhou2013active}. Secondly, the nuclear norm is a continuous function, which is important for many learning tasks. 

In this paper, we define the intrinsic reward as follows: $r^{int} = \lambda  \left\| \mathbf{Z} \right\|_\ast / \left\| \mathbf{Z} \right\|_F$,
where $\lambda $ is a weight to adjust the range of value of the nuclear norm. In the following part, we will endeavor to analyze the properties of nuclear norm to explain why divide $\left\| \mathbf{Z} \right\|_\ast$ by $\left\| \mathbf{Z} \right\|_F$, and try to obtain an adapted parameter $\lambda $ instead of using a fixed hyper-parameter. According to Cauchy–Schwarz inequality, we have:
\begin{equation}
    \left\|\mathbf{Z}\right\|_\ast = \sum_{i=1}^d \sigma_i \leq \sqrt{\sum_{i=1}^d 1 \cdot \sum_{i=1}^d \sigma_i^2 }=\sqrt{d} \sqrt{\sum_{i=1}^d \sigma_i^2} = \sqrt{d}\cdot \left\|\mathbf{Z}\right\|_F.
\end{equation}
Apparently, $\sqrt{\sum_{i=1}^d\sigma_i^2} \leq \sum_{i=1}^d \sigma_i$, 
so $\left\|\mathbf{Z}\right\|_F \leq \left\|\mathbf{Z}\right\|_\ast$. Thus, we can get the following inequality:
\begin{equation}
    \frac{1}{\sqrt{d}}\left\|\mathbf{Z}\right\|_\ast \leq \left\|\mathbf{Z}\right\|_F \leq \left\|\mathbf{Z}\right\|_\ast  \leq \sqrt{d}\cdot \left\|\mathbf{Z}\right\|_F.
    \label{eq6}  
\end{equation}

This inequality shows that $\left\|\mathbf{Z}\right\|_\ast$ and $\left\|\mathbf{Z}\right\|_F$ could bound each other. That is, if $\left\|\mathbf{Z}\right\|_\ast$ becomes larger, $\left\|\mathbf{Z}\right\|_F$ tends to be larger. Furthermore, \citef{cui2020towards} has proved that $\left\|\mathbf{Z}\right\|_F$ is strictly opposite to Shannon entropy in monotony, and maximizing $\left\|\mathbf{Z}\right\|_F$ is equal to minimizing entropy. As a result, the influence factor of $\left\|\mathbf{Z}\right\|_\ast$ could be separated into two parts: the first item is high diversity, and the second one corresponds to low entropy. Our purpose is to encourage agent to visit more novelty state, and ``diversity is what we need''. However, in state space, lower entropy means higher state aggregation, that is, higher state similarity. So we aim to promote the first effect and suppress the second item. Here, we divide $\left\| \mathbf{Z} \right\|_\ast$ by $\left\| \mathbf{Z} \right\|_F$.
\par
According to Equation \ref{eq6}, we can get: $1\le \left\|\mathbf{Z}\right\|_\ast / \left\|\mathbf{Z}\right\|_F \le \sqrt{d}$. Obviously, directly applying this scale as reward can be detrimental to the learning performance. Besides, $d$ can be changed with different environment or different architectures of neural network, so it is desirable to re-scale for $d$.
As $d \le \max(m,n)$, we get:
\begin{equation}
    1\le \frac{\left\|\mathbf{Z}\right\|_\ast }{\left\|\mathbf{Z}\right\|_F} \le \sqrt{d} \le \sqrt{\max(m,n)} ,
    \label{eq7}
\end{equation}
\begin{equation}
\frac{1}{\sqrt{\max(m,n)}}\le \frac{\left\|\mathbf{Z}\right\|_\ast }{\left\|\mathbf{Z}\right\|_F \cdot \sqrt{\max(m,n)}} \le 1.
\end{equation}
Based on the above analysis, in order to adapt the intrinsic reward to a stable range, the parameter can be set using an automated manner: $\lambda = 1/ \sqrt{\max(m,n)}$. 
Therefore, the whole intrinsic reward can be defined as:
\begin{equation}
    r^{int} = \frac{\left\|\mathbf{Z}\right\|_\ast}{\left\|\mathbf{Z}\right\|_F \cdot \sqrt{\max(m,n)}}.
\end{equation}
Finally, we can get:
\begin{equation}
    \max_{\theta_{P}} \mathbb{E}_{\pi\left(x_{t} ; \theta_{P}\right)}\left[\sum \gamma^{t} (\alpha  r_{t}^{int} + \beta r_{t}^{ext})\right],
\end{equation}
\begin{equation}
   i.e.,\quad \max_{\theta_{P}} \mathbb{E}_{\pi\left(x_{t} ; 
\theta_{P}\right)}\left[\sum \gamma^{t}\left (\alpha \frac{\left\| \mathbf{Z} \right\|_\ast} {\left(\mathbf{Z}_F\cdot \sqrt[]{\max (m,n)} \right)} + \beta r_t^{ext} \right )
    \right],
\end{equation}
where $\theta_{P}$ represents parameters of policy, $\gamma$ is the discounted factor, $\alpha$ and $\beta$ are the coefficient of intrinsic and extrinsic reward respectively. Specifically, $\alpha=1$ and $\beta=0$ refer to training policy with only intrinsic reward. Then we can use the total reward $(\alpha  r_{t}^{int} + \beta r_{t}^{ext})$ to guide the training process, and the action which obtains higher accumulated reward will be encouraged more greatly. Many classic algorithms such as PPO \citen{ppo} and DQN \citen{dqn} can be adopted as the base RL algorithm, and we use PPO \citen{ppo} in our implementation. Figure \ref{fig-whole-frame} and Algorithm \ref{alg1} present the whole framework and pseudo-code of NNM. 

\par
\begin{algorithm}[!htbp] 

\label{alg1}
 \caption{Nuclear Norm Maximization-based Intrinsic Rewards} 
    Randomly Initialize policy network $\pi_{\theta_P}$ \\
    \Repeat{max iteration or time reached}
    {    
    \For{$t=1,\cdots,T$} 
        { 
            Receive observation $s_t$ from environment \\
            $a_t \leftarrow \pi_{\theta_P}(a|s_t)$ based on policy network $\pi_{\theta_P}$  \\
            Take action $a_t$, receive observation $s_{t+1}$ and extrinsic reward $r^{ext}_t$ from environment \\  
            Generate encoded state vectors $\hat{\mathbf{z}}^1,\hat{\mathbf{z}}^2,\cdots,\hat{\mathbf{z}}$ \\    
            Form matrix $\mathbf{Z}$ by concatenating each state vector $[\hat{\mathbf{z}}^1,\hat{\mathbf{z}}^2,\cdots,\hat{\mathbf{z}}]$ \\
            Calculate intrinsic reward $r^{int}_t = \left\|\mathbf{Z}\right\|_\ast/(\left\|\mathbf{Z}\right\|_F \cdot \sqrt{\max(m,n)})$ \\
            Calculate total reward $r^{total}_t=\alpha r^{int}_t + \beta r^{ext}_t$ \\ 
            $s_t\leftarrow s_{t+1}$ \\
            Collect each step data and store episodic data when an episode is finished \\
        } 
        Sample batch data as $[(s_t, a_t, r^{total}_t),\dots,]$ from replay buffer \\
        Update $\theta_P$ with sampled data by maximizing $r^{total}_t$ using RL algorithm  \\
    }
\end{algorithm}

\section{Results}{\label{Results}}
\textbf{Experimental Settings.} We test our methods on two kinds of widely-used environments: the Atari Suite \citen{atari} and the DeepMind Control Suite \citen{dm_control}. To conduct quantitative comparisons, we follow the Disagreement~\citen{Disagreemet} settings for the Atari Suite and settings of URLB~\citen{laskin2021urlb} for the DeepMind Control Suite in our experiments.


\begin{table}[ht]
\centering
\caption{\footnotesize{Performance comparison of curiosity methods with only intrinsic reward on 26 Atari games subset.}}
\label{table-subset-Atari}
\begin{tabular}{lll|lll}
\toprule
Game           & Random   & Human   & Disagreement     & ICM             & NNM (ours)        \\
\midrule
Alien          & 227.8    & 7127.7  & 363.5            & 211.5           & \textbf{516.5}   \\
Amidar         & 5.8      & 1719.5  & 137.2            & 102.2           & \textbf{222.1}   \\
Assault        & 222.4    & 742.0   & 332.2            & \textbf{351.4}  & 343.6            \\
Asterix        & 210.0    & 8503.3  & 735.0            & 501.3           & \textbf{1395.0}  \\
Bank Heist     & 14.2     & 753.1   & \textbf{292.0}   & 282.1           & 52.3             \\
BattleZone     & 2360.0   & 37187.5 & \textbf{7230.0}  & 2380.0               & 4730.0           \\
Boxing         & 0.1      & 12.1    & 4.6              & 4.5              & \textbf{4.7}     \\
Breakout       & 1.7      & 30.5    & 55.5             & 175.5           & \textbf{164.0}   \\
ChopperCommand & 811.0    & 7387.8  & 992.0            & 582.8           & \textbf{1324.0}  \\
Crazy Climber  & 10780.5  & 23829.4 & \textbf{33954.0} & 13125.5         & 8522.0           \\
Demon Attack   & 107805.0 & 35829.4 & 142.4            & \textbf{212.4}  & 149.2            \\
Freeway        & 0.0      & 29.6    & 3.2              & 3.0             & \textbf{3.3}     \\
Frostbite      & 65.2     & 4334.7  & 239.1            & \textbf{512.5}  & 120.6            \\
Gopher         & 257.6    & 2412.5  & 2605.2           & 2506.2          & \textbf{5929.8}  \\
Hero           & 1027.0   & 30826.4 & 2942.2           & 3024.6               & \textbf{4404.0}  \\
Jamesbond      & 29.0     & 302.8   & 674.5            & 687.6               & \textbf{3053.5}  \\
Kangaroo       & 52.0     & 3035.0  & 457.0            & \textbf{751.0}  & 669.0            \\
Krull          & 1598.0   & 2665.5  & 2740.9           & 3120.5          & \textbf{4977.5}  \\
Kung Fu Master & 258.5    & 22736.3 & 4976.0           & 3987.5          & \textbf{9687.0}  \\
Ms Pacman      & 307.3    & 6951.6  & 385.7            & 448.2           & \textbf{564.1}   \\
Pong           & -20.7    & 14.6    & -3.6             & -3.5           & \textbf{-0.8}    \\
Private Eye    & 24.9     & 69571.3 & \textbf{36.0}    & -500.0          & 0.0              \\
Qbert          & 163.9    & 13455.0 & 913.8            & \textbf{2735.0} & 2041.8           \\
Road Runner    & 11.5     & 7845.0  & \textbf{1885.0}  & 1535.0          & 56.0             \\
Seaquest       & 68.4     & 42054.7 & 357.6            & 513.6           & \textbf{777.8}   \\
Up N Down      & 533.4    & 11693.2 & 11270.5          & 12271.5         & \textbf{19109.5} \\
\midrule
Mean HNS       & 0.0      & 1.0     & 0.51             & 0.50             & \textbf{1.09}             \\
\# Superhuman     & 0        & N/A    & 5                 &4                 & 5                 \\
\bottomrule
\end{tabular}
\end{table}
\textbf{NNM outperforms previous curiosity based methods.}
Table~\ref{table-subset-Atari} lists aggregate metrics and scores of three agents trained with only intrinsic reward on the Atari 26 (raw scores of the full 57 Atari games can be found in the supplementary). Human and random scores are adopted from \citef{hessel2018rainbow} and ICM scores are from \citef{burda2018large}. As done in previous works \citen{liu2021behavior,yarats2020image,SPR2020data}, we normalize the episodic return as human-normalized scores (HNS) by expert human scores to account for different score scales in each game.  \# Superhuman means the game numbers exceeding human scores.
Specifically, HNS is calculated as the average of $(\text {agent score}-\text {random score})/(\text {human score}-\text {random score})$ of all games. 
NNM displays an overwhelming superiority over Disagreement and ICM, since it surpasses both baselines in most 26 games. 
Even though NNM lost to Human in most games, it still has excellent advantages in a few games, such as Jamesbond and Breakout, thus obtaining a high HNS score of 1.09, twice of Disagreement and ICM. 
Overall, the HNS of NNM with only intrinsic reward means it nearly approximates the human level.
For the full 57 Atari games (Table \ref{table-Atari}), NNM also gets a high HNS score of 0.86, about 50\% relative improvement compared with Disagreement and ICM. Besides, NNM achieves super-human performance on nine games, slightly surpassing another two baselines. 

\begin{wraptable}{r}{0.5\textwidth}
\centering
\caption{\footnotesize{Performance comparison of curiosity methods with only intrinsic reward on full 57 Atari games.}}
\label{table-Atari}
\resizebox{\linewidth}{!}{ 
\begin{tabular}{l|ccc}
\toprule
Index                 & Disagreement &ICM & NNM(ours)      \\  
\midrule
Mean HNS               & 0.52         & 0.54        &\textbf{0.86}  \\ 
\#Superhuman             & 8            & 8           & \textbf{9}       \\ 
\bottomrule
\end{tabular}
}
\end{wraptable}

\par
\begin{figure*}[!htb]       
    \centerline{\includegraphics[width=1\textwidth]{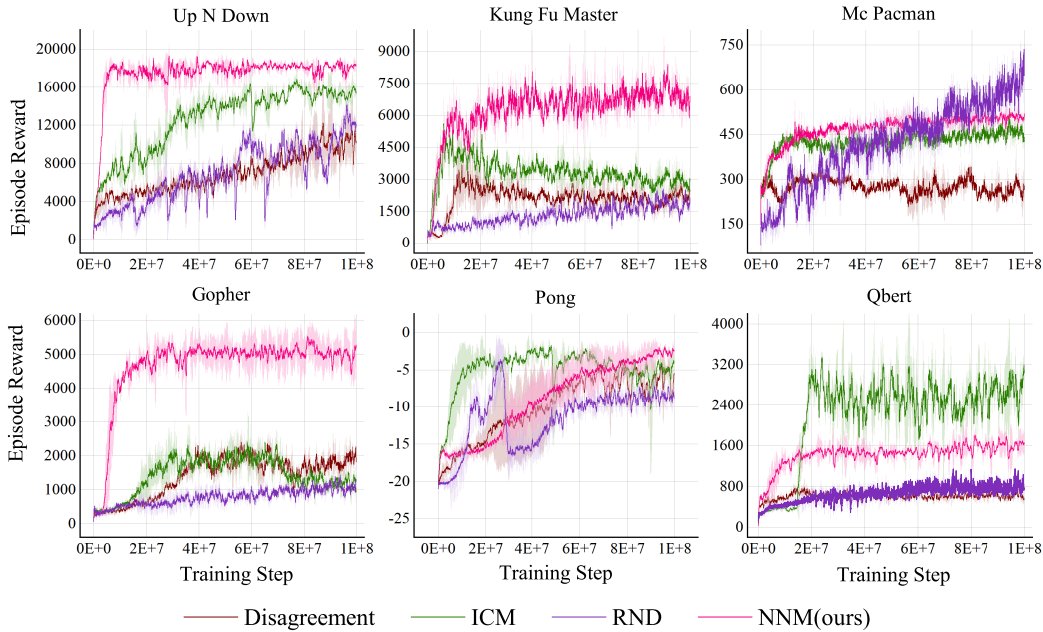}}
    \caption{\footnotesize{Performance comparison on six randomly selected Atari games with only intrinsic reward.}}
    \label{fig-atari}
\end{figure*}

\par
Figure~\ref{fig-atari} compares the learning curves of NNM with three baselines, including Disagreement~\citen{Disagreemet}, ICM~\citen{ICM} and RND~\citen{RND}, on 6 random choosed Atari games.
NNM has shown evident advantages in most games on the performance and learning speed. Especially on the Gopher and Kung Fu Master, the convergent episode reward of NNM is more than twice that of other methods. 
\par

\par
\begin{figure*}[ht] 
    \centering
    \includegraphics[width=1\textwidth]{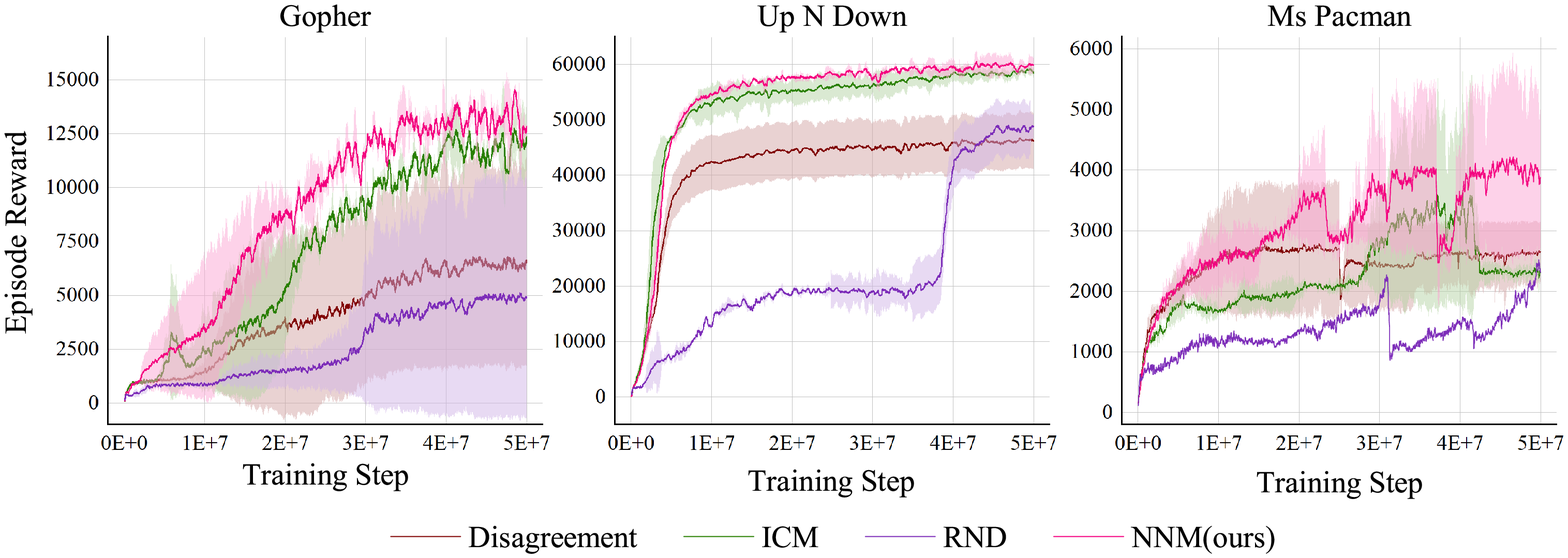}   
    \caption{\footnotesize{Performance comparison on three randomly selected Atari games, by combining the intrinsic and extrinsic rewards.}} 
    \label{fig-atari-int+ext}
\end{figure*}

We also compare the performance of agents trained with both intrinsic reward and extrinsic reward. The intrinsic and extrinsic reward coefficients are set to $\alpha=1$ and  $\beta=2$ for all methods. As shown in Figure \ref{fig-atari-int+ext}, NNM has got more significant scores and minor variance compared with other methods, proving the effectiveness of NNM further.
\par

\par
\par



\textbf{NNM outperms previous pre-training strategy}. 
To obtain better sample efficiency for DRL, many well-performed approaches employ the pre-training and fine-tune paradigm~\citen{laskin2021urlb}.
Therefore, we conduct more experiments on the DeepMind Control Suite to demonstrate the effectiveness of NNM for the pre-training. Specifically, we test Disagreement, APT and NNM on three domains~\citen{laskin2021urlb}, i.e., Walker, Quadruped and Jaco Arm (from easiest to hardest), each of which contains four tasks. In the pre-training phase, the agent is trained for 2 million steps with only intrinsic rewards. In the fine-tuning phase, the pre-trained agent learns from extrinsic rewards with 100k steps, which is far less than the pre-training. The above strategy is conducted on all 12 tasks for all methods. 

\begin{figure*}[ht] 
    \centering
    \includegraphics[width=1\textwidth]{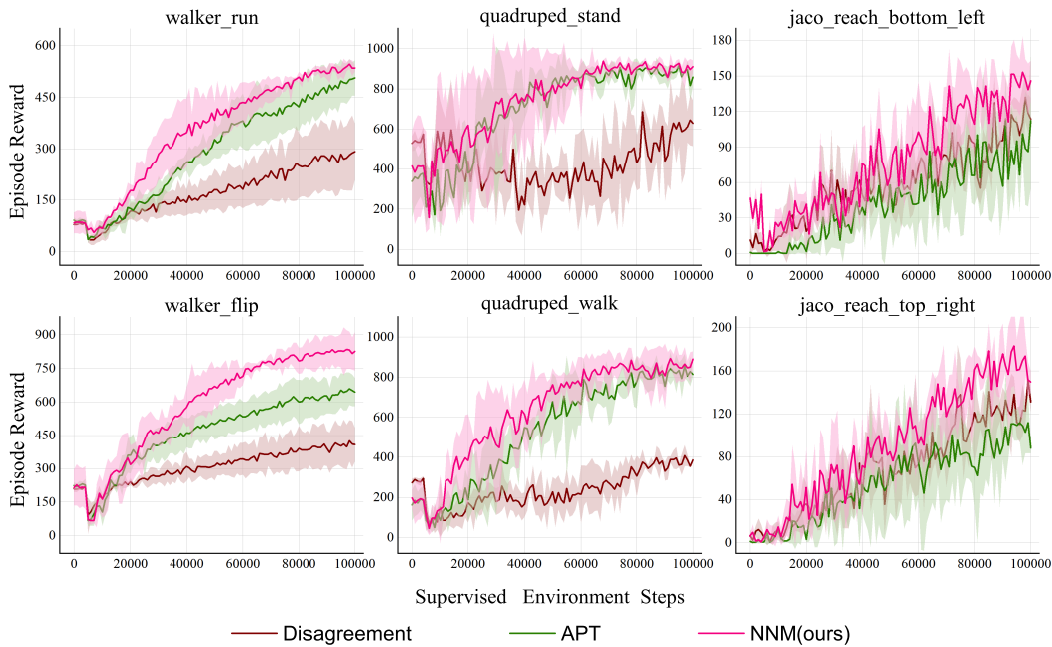}   
    \caption{\footnotesize{Performance of different methods in the fine-tuning phase.} } 
    \label{fig-DMC-res} 
\end{figure*}

Figure~\ref{fig-DMC-res} plots learning curves (fine-tuning phase) of three methods in 6 of 12 tasks. 
NNM displays a comparable convergence speed as APT, both of which are superior to Disagreement. 
However, the convergence result of NNM surpasses APT significantly. Table~\ref{table-DMC} reports the final scores of all models and their standard deviations. 
The performance of ICM, RND and APS~\citen{liu2021aps} are reported by URLB~\citen{laskin2021urlb}. The results show that NNM achieves new state-of-the-art results on 11 of the 12 tasks, indicating its ability of boosting the performance and robustness of model by pre-training.
\begin{table}[ht]
\caption{\footnotesize{Quantitative performance comparison with different curiosity methods on DMC.}}
\label{table-DMC}
\resizebox{\linewidth}{!}{    
\renewcommand{\arraystretch}{1.3} 
\begin{tabular}{lc|cccccr}
\cline{1-8}
Domain                     & Task               & ICM          & RND           & APS           & Disagreement  & APT          & NNM \\                                    \cline{1-8}
\multirow{4}{*}{Walker}    & Flip               & $417\pm16$   & $474\pm39$    & $546\pm20$    & $407\pm75$    & $653\pm62$   & \textbf{825$\pm$86}   \\ 
                          & Run                & $247\pm21$   & $406\pm30$    & $134\pm16$    & $291\pm81$    & $506\pm51$   & \textbf{535$\pm$9}     \\ 
                          & Stand              & $859\pm23$   & $911\pm5$     & $721\pm44$    & $680\pm107$   & $933\pm3$    & \textbf{961$\pm$2}     \\ 
                          & Walk               & $627\pm42$   & $704\pm30$    & $527\pm79$    & $595\pm153$   & $925\pm11$   & \textbf{949$\pm$13}        \\ \cline{1-8}
\multirow{4}{*}{Quadruped} & Jump               & $178\pm35$   & $637\pm12$    & $463\pm51$    & $383\pm265$   & $688\pm6$    & \textbf{717$\pm$30}    \\ 
                          & Run                & $110\pm18$   & $459\pm6$     & $281\pm17$    & $389\pm61$    & \textbf{498$\pm$32}   & $491\pm48$ \\ 
                          & Stand              & $312\pm68$   & $766\pm43$    & $542\pm53$    & $628\pm114$   & $858\pm44$   & \textbf{910$\pm$31}    \\
                          & Walk               & $126\pm27$   & $536\pm39$    & $436\pm79$    & $384\pm28$    & $813\pm10$   &\textbf{890$\pm$38}        \\ \cline{1-8}
\multirow{4}{*}{Jaco}      & Reach bottom left  & $111\pm11$   & $110\pm5$     & $76\pm8$      & $113\pm17$    & $112\pm57$   & \textbf{146$\pm$33}  \\ 
                          & Reach bottom right & $97\pm9$     & $117\pm7$     & $88\pm11$     & $142\pm3$     & $137\pm18$   & \textbf{171$\pm$16}    \\ 
                          & Reach top left     & $82\pm14$    & $99\pm6$      & $68\pm6$      & $121\pm17$    & $146\pm46$   & \textbf{185$\pm$3}  \\ 
                          & Reach top right    & $103\pm11$   & $100\pm6$     & $76\pm10$     & $131\pm10$    & $87\pm4$     & \textbf{149$\pm$9}        \\ \cline{1-8}
\end{tabular}
}
\end{table}

\begin{figure*}[ht] 
    \centering        \centerline{\includegraphics[width=1\textwidth]{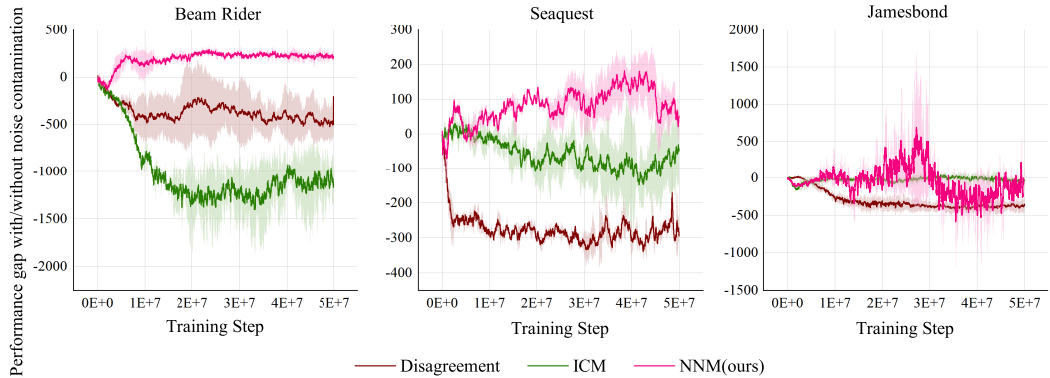}}    
    \caption{\footnotesize{The Performance gap with/without noise between Baselines and NNM (ours) on three Atari games.} } 
    \label{fig-Atari-noise} 
\end{figure*}

\textbf{NNM is robust to the noise.} 
As mentioned before, predicting the future in the face of environmental uncertainty is extremely difficult. To verify the robustness\cut{ of NNM and baselines}, we add Gaussian noise with standard deviation of 0.25 to the feature vector as the noisy \cut{version}component of the environments~\citen{dean2020see}. 
Figure~\ref{fig-Atari-noise} plots the performance gap of agents perturbed with and without noise across three environments. 
The performance of Disagreement degrades evidently in all three environments, and ICM also experiences such degradation in the former two environments. Such evidence indicates that Disagreement and ICM could overfit the noise, thus being not robust. On the contrary, the performance of NNM degrades only in the Jamesbond environment but improves instead in another two environments. This interesting phenomenon implies that NNM may leverage noise as a profitable stimulation to some extent. Overall, these comparisons strongly prove the robustness of NNM over Disagreement and ICM.



\section{Conclusion}{\label{conclusion}}
A novel intrinsic reward for DRL is proposed to address the issue of sparse extrinsic rewards, leveraging the nuclear norm maximization. Our key contribution is introducing a simple yet effective intrinsic reward derived from nuclear norm maximization, which allows the task-agnostic learning. Our method can provide high tolerance to the undesirable noise and the outliers. Extensive experiments on Atari games and DMControl suite indicate our solution can consistently improves the performance on multiple tasks. Compared with fully supervised canonical DRL algorithms, NNM achieves better or comparable results on the Atari games. The results also suggest that: NNM outperforms all previous unsupervised pre-training method for DRL. 

\setcitestyle{numbers} 
\bibliographystyle{plainnat} 
\bibliography{ref.bib} 

\end{document}